\documentclass[fleqn,10pt]{wlpeerj}
\usepackage{ulem}
\usepackage{hyperref}
\usepackage[square,sort,comma,numbers]{natbib}

\title{Probabilistically Sampled and Spectrally Clustered Plant Genotypes using Phenotypic Characteristics}

\author[1]{Aditya A. Shastri}
\author[1]{Kapil Ahuja}
\author[2]{Milind B. Ratnaparkhe}
\author[3]{Yann Busnel}
\affil[1]{Data and Computational Sciences Lab, IIT Indore, India}
\affil[2]{ICAR-Indian Institute of Soybean Research, Indore, India}
\affil[3]{Network Systems, Cybersecurity and Digital Law Department, IMT Atlantique, Rennes, France.}
\corrauthor[1]{Kapil Ahuja}{kahuja@iiti.ac.in}

\begin{abstract}
Phenotypic characteristics of a plant genotype refer to its physical properties as cataloged by plant biologists at different research centers around the world. Clustering genotypes based upon their phenotypic characteristics is used to obtain diverse sets of parents that are useful in their breeding programs.
The Hierarchical Clustering (HC) algorithm is the current standard in clustering of phenotypic data. This algorithm suffers from low accuracy and high computational complexity issues.
To address the accuracy challenge, we propose the use of Spectral Clustering (SC) algorithm. To make the algorithm computationally cheap, we propose using sampling, specifically, Pivotal Sampling that is probability based. Since application of samplings to phenotypic data has not been explored much, for effective comparison, another sampling technique called Vector Quantization (VQ) is adapted for this data as well. VQ has recently given promising results for genome data.

The novelty of our SC with Pivotal Sampling algorithm is in constructing the crucial similarity matrix for the clustering algorithm and defining probabilities for the sampling technique. Although our algorithm can be applied to any plant genotypes, we test it on the phenotypic data obtained from about 2400 Soybean genotypes. SC with Pivotal Sampling achieves substantially more accuracy (in terms of Silhouette Values) than all the other proposed competitive clustering with sampling algorithms (i.e. SC with VQ, HC with Pivotal Sampling, and HC with VQ). The complexities of our SC with Pivotal Sampling algorithm and these three variants are almost same because of the involved sampling. In addition to this, SC with Pivotal Sampling outperforms the standard HC algorithm in both accuracy and computational complexity. We experimentally show that we are up to 45\% more accurate than HC in terms of clustering accuracy. The computational complexity of our algorithm is more than a magnitude lesser than HC.
\end{abstract}

\begin{document}

\flushbottom
\maketitle
\thispagestyle{empty}

\section{Introduction}
\label{sec:1}
Variabilities present among the different plant species (also called genotypes) are useful in their breeding programs. Here, the selection of diverse parent genotypes is important. More diverse the parents, the higher are the chances of developing new plant varieties having excellent qualities \citep{cite2}. A commonly used technique here is to study the genetic variability, which looks at the different genome sequences. 
However, this kind of analysis requires a large number of sequences, while very few are available \citep{Ingvarsson2011,reseq302} because genetic sequencing is computationally and monetarily expensive \citep{sequencing}.

Variabilities in plant genotypes can also be studied using their phenotypic characteristics (physical characteristics). This kind of analysis can be relatively easily done because a sufficiently large amount of data is available from different geographical areas. In the phenotypic context, which is our first focus, a few characteristics that play an important role are Days to 50\% Flowering, Days to Maturity, Plant Height, 100 Seed Weight, Seed Yield Per Plant, Number of Branches Per Plant, etc.

Cluster analysis is an important tool to describe and summarize the variation present between different plant genotypes \citep{cite2}. Thus, clustering can be used to obtain diverse parents which, as mentioned above, is of paramount importance. It is obvious that after clustering, the genotypes present in the same cluster would have similar characteristics, while those present in different clusters would be diverse.
Phenotypic data for the genotypes of different plants (e.g., Soybean, Wheat, Rice, Maize, etc.) usually have enough variation for accurate clustering. However, if this data is obtained for the genotypes of the same plant, then clustering becomes challenging due to less variation in the data, which forms our second focus.

Hierarchical Clustering (HC) is a traditional and standard method that is currently being used by plant biologists for grouping of phenotypic data \citep{cite2,wheat,cluster-analysis-1}. However, this method has a few disadvantages. First, it does not provide the level of accuracy required for clustering similar genotypes \citep{hc}. Second, HC is based on building a hierarchical cluster tree (also called dendrogram), which becomes cumbersome and impractical to visualize when the data is too large. 

To overcome the two disadvantages of HC, in this paper, we propose the use of the Spectral Clustering (SC) algorithm. SC is mathematically sound and is known to give one of the most accurate clustering results among the existing clustering algorithms \citep{sc}. For genome data, we have recently shown substantial accuracy improvements by using SC as well \cite{vqsc}. Furthermore, unlike HC, SC does not generate the intermediate hierarchical cluster tree. To the best of our knowledge, this algorithm has not been applied to phenotypic data in any of the previous works (see the Literature Review Section below). 

HC, as well as SC, both are computationally expensive. They require substantial computational time when clustering large amounts of data \citep{sc,hc_complexity}. Hence, we use sampling to reduce this complexity. Probability-based sampling techniques have recently gained a lot of attention because of their high accuracy at reduced cost \citep{sampling}. Among these, Pivotal Sampling is most commonly used, and hence, we apply it to phenotypic data \citep{pivotal}. Like for SC, using Pivotal Sampling for phenotypic data is also new. Recently, Vector Quantization (VQ) has given promising results for genetic data \citep{vqsc}. Hence, here we adapt VQ for phenotypic data as well. This also serves as a good standard against which we compare Pivotal Sampling.

To summarize, in this paper, we develop a modified SC with Pivotal Sampling algorithm that is especially adapted for phenotypic data. The novelty of our work is in constructing the crucial similarity matrix for the clustering algorithm and defining the probabilities for the sampling technique. Although our algorithm can be applied to any plant genotypes, we test it on around 2400 Soybean genotypes obtained from Indian Institute of Soybean Research, Indore, India \citep{phenotypic-data}. In the experiments, we perform two sets of comparisons. {\it First}, our algorithm outperforms all the proposed competitive clustering algorithms with sampling in terms of the accuracy (i.e. modified SC with VQ, HC with Pivotal Sampling, and HC with VQ). The computational complexities of all these algorithms are similar because of the involved sampling. {\it Second}, our modified SC with Pivotal Sampling doubly outperforms HC, which as earlier, is a standard in the plant studies domain. In terms of the accuracy, we are up to 45\% more accurate. In terms of complexity, our algorithm is more than a magnitude cheaper than HC. 

The rest of this paper is organized as follows. Section \ref{sec:2} provides a brief summary of the previous works on HC for phenotypic data.\footnote{Since none of the previous works have used sampling for phenotypic data, we could not review this aspect.} The standard algorithms for Pivotal Sampling and SC are discussed in Section \ref{sec:3}. Section \ref{sec4} describes the crucial adaptations done in Pivotal Sampling and SC for phenotypic data. The validation metric, the experimental set-up, and the results are presented in Section \ref{sec:5}. Finally, Section \ref{sec:6} gives conclusions and future work.


\section{Literature Review}
\label{sec:2}
In this section, we present some relevant previous studies on phenotypic data and the novelty of our approach. Broadly, these studies can be classified into two categories. The first category consists of the works that identify the correlation between the different phenotypic characteristics (for example, less plant height may indicate less plant yield). The second category consists of the studies that identify the genotypes having dissimilar properties for the breeding program.

Initially, we present few works belonging to the first category. Immanuel et al. \citep{rice1} in 2011 measured nine traits of 21 Rice genotypes. Grain Yield was kept as the primary characteristic, and its correlations with all others were obtained. It was observed that several traits like Plant Height, Days to 50\% Flowering, Number of Tillers per Plant, Filled Grains per Panicle and Panicle Length were positively correlated with Grain Yield.

Divya et al. \citep{rice2} in 2015 investigated the association between Infected Leaf Area, Blast Disease Susceptibility, Number of Tillers and Grain Yield traits of two Rice genotypes. The authors concluded that Infected Leaf Area had a significant positive correlation with leaf's Blast Susceptibility. Also, Number of Tillers exhibited the highest association with Grain Yield.

Huang et al. \citep{leaf} in 2018 exploited leaf shapes and clustered 206 Soybean genotypes into three clusters using the $k$-means clustering algorithm. These clusters contained genotypes having elliptical leaves, lanceolate leaves and round leaves. Then, the authors studied variation present among the different phenotypic characteristics of these categories. They deduced that the cluster containing lanceolate leaves had maximum average Plant Height, Number of Pods per Plant, Number of Branches per Plant, and 100-Seed Weight, while the other two clusters had less values for these traits.

Carpentieri-Pipolo et al. \citep{endobacteria} in 2019 investigated the effects of endophytic bacteria on $45$ phenotypic characteristics of a Soybean genotype. The authors clustered these bacteria into three clusters using Unweighted Pair Group Method using Arithmetic Mean (UPGMA) and studied whether the bacteria had positive or negative activity on the phenotypic traits. 
It was found that five bacteria in one cluster had positive activities on almost $40$ characteristics, while fifteen bacteria in the remaining two clusters had positive activities on around 25 characteristics only.

Next, we present works that belong to the second category. Sharma et al. \citep{wheat} in 2014 performed clustering of $24$ synthetic Wheat genotypes (lines) to identify High Heat Tolerant (HHT) lines among them. Cluster analysis was performed using HC, and the Wheat lines were grouped into three clusters. This study aimed to improve heat tolerance of Wheat lines in a breeding program by identifying the diverse Wheat breeders. A similar study for accessing the drought tolerance of spring and winter bread Wheat using phenotypic data was conducted by Dodig et al. in 2010 \citep{wheat2}.

Painkra et al. \citep{cite2} in 2018 performed clustering of $273$ Soybean genotypes to determine the diversity among them. Here, the authors clustered them into seven groups using HC. Most diverse genotypes were obtained from the clusters having the highest inter-cluster distance values. According to the authors, choosing the genotypes belonging to the distant clusters maximized heterosis\footnote{Heterosis refers to the phenomenon in which a hybrid plant exhibits superiority over its parents in terms of Plant Yield or any other characteristic.} in cross-breeding.

Kahraman et al. \citep{cluster-analysis-1} in 2014 analyzed the field performances of 35 promising Common Bean genotypes. These genotypes were clustered into three groups using HC. Again, the goal of this work was to provide information for selecting the most diverse genotypes for breeding programs.  

Finally, we present a few works that belong to both the categories. Stansluos et al. \citep{corn} in 2019 analyzed $22$ phenotypic traits for $11$ Sweet Corn genotypes (cultivars). The authors showed a positive and significant correlation of Yield of Marketable Ear (YME) with Ear Diameter (ED) and Number of Marketable Ear (NME), whereas YME was negatively correlated with Thousand Kernel Weight (TKW). Also, they grouped these cultivars into four clusters using HC to obtain variation among them.

Fried et al. \citep{root} in 2018 determined whether the root traits were related to other phenotypic traits for 49 Soybean genotypes. In this work, Principal Component Analysis (PCA) biplot was used to separate the Soybean genotypes into different clusters. According to the authors, this research was critical for Soybean improvement programs since it helped select genotypes with the improved root characteristics. 

We have three contributions as below, which have not been catered in any of the above cited papers.
\begin{enumerate}[noitemsep]
	\item With a focus on the second category, we perform grouping of several thousand genotypes as compared to a few hundred in the papers cited above.
	\item Clustering becomes computationally expensive when the size of the data is very large. Hence, sampling is required to make the underlying algorithm scalable. As mentioned in the previous section, we adapt Pivotal Sampling, a probability-based technique, for phenotypic data.
	\item HC, which is the most common clustering algorithm (and some other sporadically used algorithms like $k$-means and UPGMA), do not provide the level of accuracy needed. Again, as earlier, we develop a variant of the SC algorithm, which is considered highly accurate, especially for phenotypic data.
	
\end{enumerate}

\section{Sampling and Clustering Algorithms}
\label{sec:3}
In this section, we briefly discuss the standard algorithms for Pivotal Sampling and SC in the two subsections below.

\subsection{Pivotal Sampling}
\label{sec:3.1}
This is a well-developed sampling theory that handles complex data with unequal probabilities. The method is attractive because it can be easily implemented by a sequential procedure, i.e. by a single scan of the data \citep{pivotal2}. Thus, the complexity of this method is $\mathcal{O}(n)$, where $n$ is the population size. It is important to emphasize that the method is independent of the density of the data. 

Consider a finite population $U$ of size $n$ with its each unit identified by a label $i = 1, 2, ..., n$. A sample $S$ is a subset of $U$ with its size, either being random ($N(S)$) or fixed ($N$). Obtaining the inclusion probabilities of all the units in the population, denoted by $\pi_i$ with $i=1, 2, ..., n$, forms an important aspect of this unequal probability sampling technique.

The pivotal method is based on a principle of contests between units \citep{sampling}. At each step of the method, two units compete to get selected (or rejected). Consider unit $i$ with probability  $\pi_i$ and unit $j$ with probability  $\pi_j$, then we have the two cases as below.

\begin{enumerate}
	\item \textbf{Selection step ($\pi_i+\pi_j\geq1$):} Here, one of the units is selected, while the other one gets the residual probability $\pi_i+\pi_j-1$ and competes with another unit at the next step. More precisely, if $(\pi_i,\pi_j)$ denotes the selection probabilities of the two units, then
	
	\begin{equation*}
	    (\pi_i,\pi_j)=\left\{\begin{matrix}
	(1,\pi_i+\pi_j-1) \textnormal{ with probability } \frac{1-\pi_j}{2-\pi_i-\pi_j}\\ 
	(\pi_i+\pi_j-1,1) \textnormal{ with probability } \frac{1-\pi_i}{2-\pi_i-\pi_j} 
	\end{matrix}\right.
	\end{equation*}
	
	\item \textbf{Rejection step ($\pi_i+\pi_j<1$):} Here, one of the units is definitely rejected (i.e. not selected in the sample), while the other one gets the sum of the inclusion probabilities of both the units and competes with another unit at the next step. More precisely,
	
	\begin{equation*}
	    (\pi_i,\pi_j)=\left\{\begin{matrix}
	(0,\pi_i+\pi_j) & \textnormal{with probability } \frac{\pi_j}{\pi_i+\pi_j}\\ 
	
	(\pi_i+\pi_j, 0) & \textnormal{with probability } \frac{\pi_i}{\pi_i+\pi_j} 
	\end{matrix}\right.
	\end{equation*}
	
\end{enumerate}

This step is repeated for all the units present in the population until we get the sample of size $N(S)$ or $N$. The worst-case occurs when we obtain the last sample (i.e. $N^{th}$ sample) in the last iteration. 

\subsection{Spectral Clustering}
\label{sec:3.2}
Clustering is one of the most widely used techniques for exploratory data analysis with applications ranging from statistics, computer science, and biology to social sciences and psychology etc. It is used to get a first impression of data by trying to identify groups having ``similar behavior" among them. Compared to the traditional algorithms such as $k$-means, SC has many fundamental advantages. Results obtained by SC are often more accurate than the traditional approaches. It is simple to execute and can be efficiently implemented by using the standard linear algebra methods. The algorithm consists of four steps as below \citep{sc}.

\begin{enumerate}
	\item The first step in the SC algorithm is the construction of a matrix called the similarity matrix. Building this matrix is the most important aspect of this algorithm; better its quality, better the clustering accuracy \citep{sc}. This matrix captures the local neighborhood relationships between the data points via similarity graphs and is usually built in three ways. The first such graph is a $\epsilon$-neighborhood graph, where all the vertices whose pairwise distances are smaller than $\epsilon$ are connected. The second is a $k$-nearest neighborhood graph, where the goal is to connect vertex $v_i$ with vertex $v_j$ if $v_j$ is among the $k$-nearest neighbors of $v_i$. The third and the final is the fully connected graph, where each vertex is connected with all the other vertices. Similarities are obtained only between the connected vertices. Thus, similarity matrices obtained by the first two graphs are usually sparse, while the fully connected graph yields a dense matrix. 
	
	Let the $n$ vertices of a similarity graph be represented numerically by vectors $a_1, a_2, ..., a_n$, respectively. Here, each $a_i$ $\in \mathbb{R}^m$ is a column vector for $i=1, ..., n$. Also, let $a_i^l$ and $a_j^l$ denote the $l^{th}$ elements of vectors $a_i$ and $a_j$, respectively, with $l=1, ..., m$. There exist many distance measures to build the similarity matrix \citep{distance}. We describe some common ones below using the above introduced terminologies.
	\begin{enumerate}[noitemsep]
		\item \textbf{City block distance:} \citep{distance} It is the special case of the Minkowski distance
		
		\begin{equation*}
		    d_{ij}=\sqrt[p]{\sum_{l=1}^{m}|a_i^l-a_j^l|^p}
		\end{equation*}
		with $p=1$.
				
		\item \textbf{Euclidean distance:} \citep{distance} It is the ordinary straight line distance between two points in the Euclidean space. It is again the special case of the Minkowski distance, where the value of $p$ is taken as $2$. Thus, it is given by
		
		\begin{equation*}
		    d_{ij}=\sqrt{\sum_{l=1}^{m}(a_i^l-a_j^l)^2}.
		\end{equation*}
		
		\item \textbf{Squared Euclidean distance:} \citep{distance} It is the square of the Euclidean distance, and is given by
		
		\begin{equation*}
		    d_{ij}=\sum_{l=1}^{m}(a_i^l-a_j^l)^2.
		\end{equation*}
		
		\item \textbf{Cosine distance:} \citep{distance} It measures the cosine of the angle between two non-zero vectors, and is given by
		
		\begin{equation*}
		    d_{ij}=1-{a_i \cdot a_j \over \|a_i\| \|a_j\|},
		\end{equation*}
		where, $\|\cdot \|$ denotes the Euclidean norm of a vector.
		
		\item \textbf{Correlation distance:} \citep{correlation} It captures the correlation between two non-zero vectors, and is given by
		
		\begin{equation*}
		    d_{ij}=1-{(a_i-\bar{a}_i)^t(a_j-\bar{a}_j) \over \sqrt{(a_i-\bar{a}_i)^t(a_i-\bar{a}_i)}\sqrt{(a_j-\bar{a}_j)^t(a_j-\bar{a}_j)}},
		\end{equation*}
		where, $\bar{a}_i$ and $\bar{a}_j$ are the means of $a_i$ and $a_j$ multiplied with a vector of ones, respectively, and $t$ signifies the transpose operation.
		
		\item \textbf{Hamming distance:} \citep{hamming} It measures the number of positions at which the corresponding values of two vectors are different, and is given by
		
		\begin{equation*}
		    d_{ij}={\#(a_i^l \neq a_j^l) \over n},
		\end{equation*}

		\item \textbf{Jaccard distance:} \cite{jaccard} It again measures the number of positions at which the corresponding values of two vectors are different excluding the positions where both the vectors have zero values, and is given by
		
		\begin{equation*}
	    d_{ij}={\#[(a_i^l \neq a_j^l)\cap ((a_i^l\neq 0)\cup (a_j^l\neq 0))] \over \#[(a_i^l\neq 0)\cup (a_j^l\neq 0)]}.
	    \end{equation*}
		
	\end{enumerate}

	\item Next, a matrix called the Laplacian matrix is constructed. This matrix is either non-normalized or normalized. The non-normalized Laplacian matrix is defined as
	
	\begin{equation*}
	L=D-W,    
	\end{equation*}
	where $W$ is the similarity matrix and $D$ is a diagonal matrix whose elements are obtained by adding together the elements of all the columns for every row of $W$.
	
	Normalized Laplacian matrix is again of two types: the symmetric Laplacian ($L_{sym}$) and the random walk Laplacian ($L_{rw}$). Both these matrices are closely related to each other and are defined as
	
	\begin{equation*}
	    L_{sym}=D^{-1/2}LD^{-1/2}=I-D^{-1/2}WD^{-1/2}.
	\end{equation*}
	\begin{equation*}
	    L_{rw}=D^{-1}L=I-D^{-1}W.
	\end{equation*}

	Henceforth, the non-normalized Laplacian matrix is referred to as the Type-1 Laplacian, $L_{sym}$ as the Type-2 Laplacian, and $L_{rw}$ as the Type-3 Laplacian. In the literature, it is suggested to use the normalized Laplacian matrix instead of the non-normalized one, and specifically the Type-3 Laplacian \citep{sc}.
	\item Once we have the Laplacian matrix, we obtain the first $k$ eigenvectors $u_1, ..., u_k$ of this matrix, where $k$ is the number of clusters.
	\item Finally, these eigenvectors are clustered using the $k$-means clustering algorithm.
\end{enumerate}

\section{Implementing Pivotal Sampling and Modified Spectral Clustering for Phenotypic Data}
\label{sec4}
Here, we first present the application of Pivotal Sampling to obtain the samples from phenotypic data. Subsequently, we implement our modified SC algorithm on the same data. Consider that the phenotypic data of a plant consist of $n$ genotypes with each genotype evaluated for $m$ different characteristics/ traits.
As discussed in Section \ref{sec:3.1}, Pivotal Sampling requires that the inclusion probabilities (i.e. $\pi_i$ for $i=1, ..., n$), of all the genotypes in the population $U$, be computed before a unit is considered for a contest. The set of characteristics associated with a genotype can be exploited in computing these probabilities. To select a sample of size $N$, where $N \ll n$, we obtain these probabilities as \citep{pivotal2}

\begin{equation}
\pi_i=N\frac{\varkappa_i}{\sum_{i\in U}\varkappa_i}, \label{eq:1}
\end{equation}
where $\varkappa_i$ can be a property associated with any one characteristic (or a combination of them) of the $i^{th}$ genotype. Obtaining $\pi_i$ in such a way also ensures that $\sum_{i=1}^{n}\pi_i=N$, i.e. we get exactly $N$ selection steps, and in-turn, exactly $N$ samples. 

In our implementation, we use the deviation property of the genotypes, which is discussed next. 
Since different characteristics have values in different ranges, we start by normalizing them as below \citep{normalize,normalize1}.

\begin{equation*}
    (\mathcal{X}_j)_i={(x_j)_i-\textnormal{min}(x_j) \over \textnormal{max}(x_j)-\textnormal{min}(x_j)}.
\end{equation*}
Here, $(\mathcal{X}_j)_i$ and $(x_j)_i$ are the normalized value and the actual value of the $j^{th}$ characteristic for the $i^{th}$ genotype, respectively with $j=1, ..., m$ and $i=1, ..., n$. Furthermore, max($x_j$) and min($x_j$) are the maximum and the minimum values of the $j^{th}$ characteristic among all the genotypes. Now, the deviation for the $i^{th}$ genotype is calculated using the above normalized values as

\begin{equation*}
    dev_i=\sum_{j=1}^{m}\textnormal{max}(\mathcal{X}_j)-(\mathcal{X}_j)_i.
\end{equation*}
Here, $\textnormal{max}(\mathcal{X}_j)$ denotes the maximum normalized value of the $j^{th}$ characteristic among all the genotypes. Practically, a relatively large value of $dev_i$ indicates that the $i^{th}$ genotype is less important, and hence, its probability should be small. Thus, the inclusion probability of a genotype is calculated by taking $\varkappa_i = \frac{1}{dev_i}$ in Eq. (\ref{eq:1}) or 

\begin{equation*}
    \pi_i=N\frac{\frac{1}{dev_i}}{\sum_{i\in U}\frac{1}{dev_i}}.
\end{equation*}
Once these probabilities are obtained, we follow the two steps (selection and rejection) as discussed in Section \ref{sec:3.1}. 

Next, we discuss the clustering of these $N$ genotypes into $k$ clusters. Similar to the standard SC algorithm discussed in Section \ref{sec:3.2}, the first step in our modified SC is to obtain the similarity matrix. As mentioned earlier, this is the most important aspect of this algorithm since better the quality of this matrix, better is the clustering accuracy.
For this, we consider these $N$ genotypes as the vertices of a graph. Let vector $a_i$ contain the normalized values of all the characteristics ($m$) for the $i^{th}$ genotype. Thus, we have $N$ such vectors corresponding to the $N$ genotypes selected using Pivotal Sampling, and each vector is of size $m$. In our implementation, we use a fully connected graph to build the similarity matrix, i.e. we obtain similarities among all the $N$ genotypes.

We define the similarity between the vectors $a_i$ and $a_j$ (representing the genotypes $i$ and $j$, respectively) as the inverse of the distance between these vectors obtained by using the distance measures mentioned in Section \ref{sec:3.2}. This is intuitive because smaller the distance between any two genotypes, larger the similarity between them and vice versa. We denote this distance by $d_{ij}$. We build this matrix of size $N \times N$ by obtaining the similarities among all the $N$ genotypes. 

The next step is to compute the Laplacian matrix, which when obtained from the above-discussed similarity matrix, generates poor eigenvalues,\footnote{Zero/ close to zero and distinct eigenvalues are considered to be a good indicator of the connected components in a similarity matrix. Thus, eigenvalues are considered poor when they are not zero/ not close to zero or indistinct \citep{sc}.} and in-turn poor corresponding eigenvectors that are required for clustering\footnote{For some distance matrices (like Euclidean distance), the eigenvalues don't even converge.}. Thus, instead of taking only the inverse of $d_{ij}$, we also take its exponent, i.e. we define the similarity between the $i^{th}$ and the $j^{th}$ genotypes as $e^{-d_{ij}}$ \citep{NgSC,spsc}. This, besides fixing the poor eigenvalues/ eigenvectors problem, also helps perform better clustering of the given data. Further, we follow the remaining steps as discussed in Section \ref{sec:3.2}.

Above, we discussed the clustering of $N$ sampled genotypes into $k$ clusters. However, our goal is to cluster all $n$ genotypes and not just $N$. Hence, there is a need to reverse-map the remaining $n-N$ genotypes to these $k$ clusters. 
For this, we define the notion of average similarity, which between the non-clustered genotype $p_i$ and the cluster $C_l$ is given as

\begin{equation*}
    \mathcal{AS}(C_l, p_i)=\frac{1}{\#(C_l)}\sum_{q\in C_l} e^{-d_{p_iq}}.
\end{equation*}

Here, $\#(C_l)$ denotes the number of genotypes present in $C_l$ and $q$ is a genotype originally clustered in $C_l$ by our modified SC algorithm with Pivotal Sampling. We obtain the average similarity of $p_i$ with all the $k$ clusters (i.e. with $C_l$ for $l=1, ..., k$), and associate it with the cluster with which $p_i$ has the maximum similarity.

Next, we perform the complexity analysis of our algorithm. Since Pivotal Sampling and SC form the bases of our algorithm, we discuss the complexities of these algorithms before ours.
\begin{enumerate}[noitemsep]
	\item Pivotal Sampling ($n$: number of genotypes, $N$: sample size)
	\begin{enumerate}[noitemsep]
		\item Obtaining Probabilities: $\mathcal{O}(n)$
		\item Obtaining Samples: $\mathcal{O}(n)$
	\end{enumerate}
	\item SC ($n$, $m$: number of characteristics)
	\begin{enumerate}[noitemsep]
		\item Constructing Similarity Matrix: $\mathcal{O}(n^2m)$
		\item Obtaining Laplacian Matrix: $\mathcal{O}(n^3)$
	\end{enumerate}
	\item Our Algorithm ($n$, $N$, $m$)
	\begin{enumerate}[noitemsep]
		\item Obtaining Samples: $\mathcal{O}(n)$
		\item Constructing Similarity Matrix: $\mathcal{O}(N^2m)$
		\item Obtaining Laplacian Matrix: $\mathcal{O}(N^3)$
		\item Reverse Mapping: $\mathcal{O}\big((n-N)N\big)$
	\end{enumerate}
	Thus, the overall complexity of our algorithm is $\mathcal{O}(nN+N^3+Nm^2)$. Here, we have kept three terms because any of these can dominate (here, $n \gg N, m$).
\end{enumerate}
When we compare complexity of our algorithm with that of HC, which is $\mathcal{O}(n^3)$, it is evident that we are more than a magnitude faster than HC.

\section{Results}
\label{sec:5}
In this section, we first briefly discuss the data used for our experiments. Next, we check the goodness of our sampling technique by estimating a measure called the population total. Subsequently, we describe the clustering set-up, where the validation metric, the ideal number of clusters, the suitable distance measures for building similarity matrices, and the most useful Laplacian matrix are discussed. Finally, we present the results for our modified SC with Pivotal Sampling. Here, we compare our algorithm with (a) SC with VQ, HC with Pivotal Sampling, HC with VQ and (b) non-sampled HC.



\subsection{Data Description}
\label{sec:5.1}
As mentioned in Introduction, our techniques can be applied to any plant data. However, here we experiment on phenotypic data of Soybean genotypes. This data is taken from Indian Institute of Soybean Research, Indore, India, and consists of $29$ different characteristics/ traits for $2376$ Soybean genotypes \citep{phenotypic-data}. Among these, we consider the following eight characteristics that are most important for higher yield: Early Plant Vigor (EPV), Plant Height (PH), Number of Primary Branches (NPB), Lodging Score (LS), Number of Pods Per Plant (NPPP), 100 Seed Weight (SW), Seed Yield Per Plant (SYPP) and Days to Pod Initiation (DPI). Table \ref{data} provides a snapshot of this data for a few genotypes.

\begin{table}[ht]
	\centering
\begin{tabular}{|c|c|c|c|c|c|c|c|c|}
\hline
			\textbf{Genotypes} & \textbf{EPV}  & \textbf{PH}  & \textbf{NPB}  & \textbf{LS}  & \textbf{NPPP}  & \textbf{SW}  & \textbf{SYPP}  & \textbf{DPI}  \\ \hline
			
			1&Poor		&54		&6.8	&Moderate	&59.8	&6.5	&2.5	&65 \\
			2&Poor		&67		&3.4	&Severe		&33		&6.2	&3.9	&64 \\
			3&Poor		&38.4	&2.8	&Slight		&68		&6.9	&4.4	&61 \\
			4&Good		&60.8	&4		&Moderate	&34.6	&6.1	&3		&65 \\
			\vdots&\vdots	&\vdots	&\vdots	&\vdots		&\vdots	&\vdots	&\vdots	&\vdots	\\
			2376&Very Good&89.6	&5		&Severe		&32.6	&7.3	&3.4	&62 \\ \hline
\end{tabular}
\caption{\label{data}Description of Phenotypic Data.}	
\end{table}

\subsection{Sampling Discussion}
\label{sec:5.2}
To inspect the quality of our sampling techniques, we estimate a measure called the population total, which is the addition of values of a particular characteristic for all the $n$ units (genotypes here) present in the population $U$. For example, if ``Plant Height (PH)" is the characteristic of interest, then the population total is the addition of PH values for all the $n$ genotypes. Mathematically, the exact (or actual) population total for a characteristic of interest $x_j$ is given as

\begin{equation*}
  Y=\sum_{i\in U}(x_j)_i,  
\end{equation*}
where, as earlier, $(x_j)_i$ is the value of the $j^{th}$ characteristic for the $i^{th}$ genotype and $U$ is the set of all genotypes. As evident by the above equation, this measure can be only applied to characteristics that contain numerical values.\footnote{Methods do exist to convert non-numeric data to numeric one for using this measure.}

In this work, we use two different estimators to compute an approximation of the population total from the sampled data. Closer the value of an estimator to the actual value, better the sampling. First is the Horvitz-Thompson (HT)-estimator (also called $\pi$-estimator), which is defined as \citep{HT}

\begin{equation*}
    Y_{HT}^{'} = Y_{\pi}^{'}=\sum_{i\in S}\frac{(x_j)_i}{\pi_i},
\end{equation*}
where, $\pi_i$ is the inclusion probability of the $i^{th}$ genotype as evaluated in Section \ref{sec4} and $S$ is the set of sampled genotypes.
Another estimator that we use is the H\'ajek-estimator. It is usually considered better than the HT-estimator and is given as \citep{hajek}

\begin{equation*}
    Y_{H\acute{a}jek}^{'} =n\frac{\sum_{i\in S}\frac{(x_j)_i}{\pi_i}}{\sum_{i\in S}\frac{1}{\pi_i}},
\end{equation*}
here, as earlier, $n$ is the total number of genotypes.

The actual population total and the values of the above two estimators for six characteristics (that have numerical values) when using Pivotal Sampling and $500$ samples are given in Table \ref{estimator} (see columns 3, 4, and 6, respectively). From this table, it is evident that the approximate values of the population total are very close to the corresponding actual values. Thus, Pivotal Sampling works well in an absolute sense. Here, we also compute the values of the two estimators when using VQ (see columns 5 and 7). We can notice from these results that VQ also works reasonably well, but Pivotal Sampling is better.

\begin{table}[ht]
\centering
\begin{tabular}{|c|c|c|c|c|c|c|}
\hline
			\textbf{Sr.}	& \textbf{Characteristics}	& \textbf{Actual}		& \textbf{Pivotal}	& \textbf{VQ} & \textbf{Pivotal}	& \textbf{VQ} 	   \\
			\textbf{No.}	& 		& \textbf{Population} 	& \textbf{Sampling} 	& \textbf{(HT)} & \textbf{Sampling} &\textbf{(H\'ajek)}	  	\\
			& 				&  	\textbf{Total}		& \textbf{(HT)} & & \textbf{(H\'ajek)}	& 	\\ \hline
			1&	PH	&121773.05	&122507.84	&123407.80	&123716.09	&113168.90 \\

			2	&NPB 	&8576.56	&8585.28	&9669.29	&8669.95	&8867.05 \\

			3	& NPPP	&99712.72	&100193.53	&114465.66	&101181.70	&104968.67 \\

			4&	SW 	&20073.32	&19907.10	&20966.86	&20103.44	&19227.28	\\

			5	&SYPP 	&10048.04	&10137.57	&10536.08	&10237.55	&9661.92 \\

			6&	DPI 	&136810	&135309.78	&149242.17	&136644.29	&136859.84 \\	\hline
			
	\end{tabular}
	\caption{\label{estimator}HT and H\'ajek estimators values for Pivotal Sampling and VQ as compared to the actual population total with $N=500$ as the sample size.}
\end{table}

\subsection{Clustering Setup}
\label{sec:5.3}
Here, {\it first}, we describe the criteria used to check the goodness of generated clusters. There are various metrics available for the validation of clustering algorithms. These include Cluster Accuracy (CA), Normalized Mutual Information (NMI), Adjusted Rand Index (ARI), Compactness (CP), Separation (SP), Davis-Bouldin Index (DB), and Silhouette Value \citep{validation,silhouette}. For using the first three metrics, we should have a prior knowledge of the cluster labels. However, here we do not have this information. Hence, we cannot use these validation metrics. Rest of the techniques do not have this requirement, and hence, can be used for validation here. We use Silhouette Value because of its popularity \citep{silhouette}.

Silhouette Value is a measure of how similar an object is to its own cluster (intra-cluster similarity) compared with other clusters (inter-cluster similarity). For any cluster $C_l$ ($l = 1, ...,  k; \textnormal{ say } l = 1$), let $a(i)$ be the average distance between the $i^{th}$ data point and all other points in the cluster $C_1$, and let $b(i)$ be the average distance between this $i^{th}$ data point in the cluster $C_1$ and all other points in clusters $C_2, ..., C_k$. Silhouette Value for the $i^{th}$ data point is defined as \citep{silhouette}

\begin{equation}
s(i) = {b(i)-a(i) \over \text{max}\{a(i), b(i)\}}, \label{eq:2}
\end{equation}
where, $a(i)$ and $b(i)$ signify the intra-cluster and the inter-cluster similarities, respectively. Silhouette Value comes to be between $-1$ and $1$,\footnote{This is because the denominator of Eq. (\ref{eq:2}) is always greater than its numerator.} and average over all the data points is computed. A positive value (tending towards 1) indicates good clustering (compact and well-separated clusters), while a negative value (tending towards -1) indicates poor clustering.

{\it Second}, we determine the ideal number of clusters by using the eigenvalue gap heuristic \citep{sc,eigenheuristic}. If $\lambda_1, \lambda_2, ..., \lambda_n$  are the eigenvalues of the matrix used for clustering (e.g., the Laplacian matrix), then often the initial set of eigenvalues, say $k$, have a considerable difference between the consecutive ones in this set. That is, $|\lambda_i - \lambda_{i+1}| \not \approx 0$ for $i=1, ..., k-1$. After the $k^{th}$ eigenvalue, this difference is usually approximately zero. According to this heuristic, this $k$ gives a good estimate of the ideal number of clusters.

\begin{figure}[ht]
\centering
\includegraphics[width=\linewidth,height=6cm]{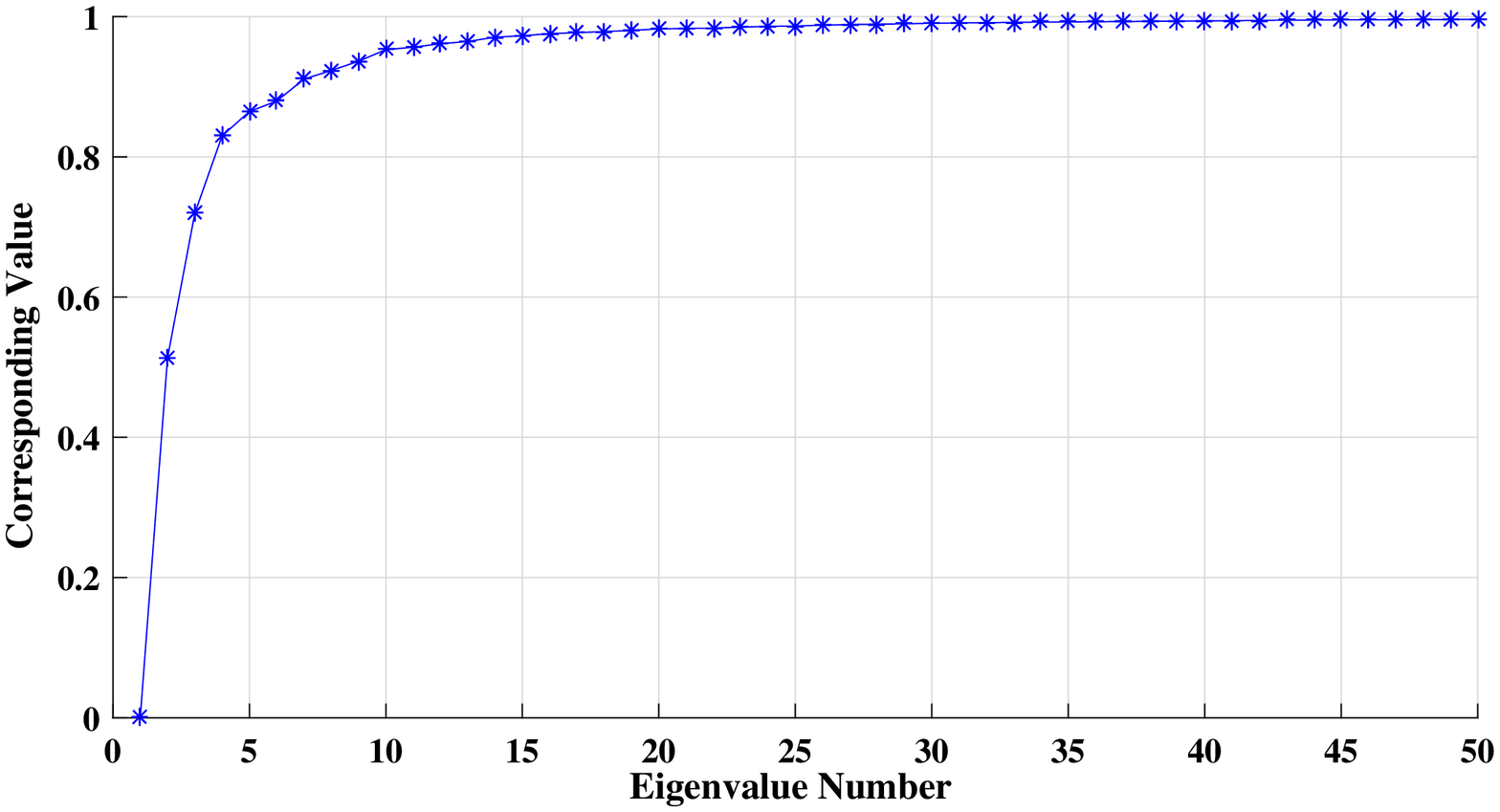}
	\caption{Fifty Smallest Eigenvalues of the Type-3 Laplacian Matrix Obtained from the Euclidean Similarity Matrix (for estimating the ideal number of clusters).}
	\label{Fig1}
\end{figure}

For this experiment, without loss of generality, we build the similarity matrix using the Euclidean distance measure on the above discussed phenotypic data.
As mentioned earlier, it is recommended to use the Type-3 Laplacian matrix \citep{sc}. Hence, we use its eigenvalues for estimating $k$.
Figure \ref{Fig1} represents the graph of the first fifty smallest eigenvalues (in absolute terms) of this Laplacian matrix. On the $x$-axis, we have the eigenvalue number, and on the $y$-axis its corresponding value.

From this figure, we can see that there is a considerable difference between the first ten consecutive eigenvalues. After the tenth eigenvalue, this difference is very small (tending to zero). Hence, based upon the earlier argument and this plot, we take $k$ as ten. To corroborate this choice more, we experiment with $k$ as twenty and thirty as well. As expected, and discussed in detail later in this section, Silhouette Values for these numbers of clusters are substantially lesser than those for ten clusters.

{\it Third}, and final, we perform experiments to identify the suitable similarity measures to build the similarity matrix, and also verify that, as recommended, the Type-3 Laplacian matrix is the best. Table \ref{suitable_similarity} below gives Silhouette Values of our modified SC for all seven similarity measures and three Laplacians when clustering the earlier presented phenotypic data into $10$, $20$, and $30$ clusters.

\begin{table}[ht]
	\centering
	\begin{tabular}{|c|c|c|c|c|c|}
	\hline
	\textbf{Sr.}	& \textbf{Similarity}	& \textbf{Number of}		& \textbf{Type-1}	& \textbf{Type-2}  	& \textbf{Type-3}  	   	\\
	\textbf{No.}	& \textbf{Measure}		& \textbf{Clusters $(k)$} 	& \textbf{Laplacian} 	& \textbf{Laplacian} 	& \textbf{Laplacian} \\ \hline
	
	1.	&Euclidean		&10		&0.0828	&-0.0273&	\textbf{0.2422} \\
		&				&20		&0.0455	&-0.1096&	\textbf{0.2069} \\
		&				&30		&0.0887	&-0.1536&	\textbf{0.1783} \\ \hline	
		
	2.	&Squared		&10		&0.0815	&-0.0555&	\textbf{0.3836} \\
		&Euclidean		&20		&-0.0315&-0.1809&	\textbf{0.2612} \\
		&				&30		&0.0354	&-0.2367&	\textbf{0.1538} \\ \hline
		
	3.	&City-block		&10		&0.0687	&0.2375	&\textbf{0.2647} \\
		&				&20		&-0.0356&	0.1347&	\textbf{0.2082} \\
		&				&30		&-0.0870&	0.0866&	\textbf{0.1887} \\ \hline	
		
	4.	&Cosine			&10		&0.1737 &-0.1408&	0.0694	 \\
		&				&20		&0.0359	&-0.1973&	0.0277	 \\
		&				&30		&0.0245	&-0.2456&	-0.0316	 \\	\hline			
		
	5.	&Correlation	&10		&0.1926&-0.1259	&\textbf{0.3426}	 \\
		&				&20		&0.0970	&-0.2198&	\textbf{0.2313}	 \\
		&				&30		&0.2383	&-0.2604&	\textbf{0.1556}	 \\	\hline
		
	6.	&Hamming		&10		&0.0643	&0.0706&0.0775	 \\
		&				&20		&0.0683	&0.0311	&0.0382	 \\
		&				&30		&0.0715	&0.0283	&0.0229	 \\	\hline	
		
	7.	&Jaccard		&10		&0.0716	&0.0303	&0.0458	\\
		&				&20		&0.0446	&0.0276	&0.0236	 \\
		&				&30		&0.0279	&0.0298	&0.0318	 \\	\hline	

	\end{tabular}
	\caption{\label{suitable_similarity}Silhouette Values for modified SC with seven similarity measures and three Laplacian matrices for $k=10, 20$, and $30$. Silhouette Values in bold represent good clustering.}
\end{table}

From this table, it is evident that Silhouette Values for the Euclidean, Squared Euclidean, City-block and Correlation similarity measures and the Type-3 Laplacian matrix are the best. Hence, we use these four similarity measures and this Laplacian matrix. Also, as mentioned earlier, Silhouette Values decrease for twenty and thirty cluster sizes.

\subsection{Results}
\label{sec:5.4}
Using the earlier presented dataset, and sampling-clustering setups, we compare our proposed algorithm (modified SC with Pivotal Sampling) with the existing variants in three ways. Initially, we compare with modified SC with VQ, HC\footnote{HC also requires building a similarity matrix.} with Pivotal Sampling and HC with VQ for a sample size of $500$.
Since the results for modified SC with VQ come out to be closest to our algorithm, next, for broader appeal we compare these two algorithms for a sample size of $300$.
Finally, we compare our algorithm with the current best in literature for this kind of data (i.e. HC without sampling) for both the sample sizes of $500$ and $300$.

The results for the {\it initial} set of comparisons are given in Table \ref{sampling500}. Columns 2 and 3 give the similarity measures and the number of clusters chosen, respectively. Columns 4 and 5 give Silhouette Values of modified SC with Pivotal Sampling and VQ, respectively, while columns 6 and 7 give Silhouette Values of HC with Pivotal Sampling and VQ, respectively.
\begin{table}[ht]
\centering
\begin{tabular}{|c|c|c|c|c|c|c|}
\hline
			\textbf{Sr.}& \textbf{Similarity} &\textbf{\# of} & \multicolumn{2}{|c|}{\textbf{modified SC}} & \multicolumn{2}{|c|}{\textbf{HC}} \\
			\cline{4-7}

			\textbf{No.}	& \textbf{Measure}	& 	\textbf{Clusters}	& \textbf{Pivotal}	& \textbf{VQ}  & \textbf{Pivotal} & \textbf{VQ}	   \\ 
			
				& 		&  $(k)$	& \textbf{Sampling} 	&  & \textbf{Sampling}  & 	\\ \hline

	1.	&Euclidean	&10		&{\bf 0.2152} &0.2061 &0.2105 &-0.1040\\
		&			&20		&{\bf 0.1905} &0.1448 &$0.2263^*$ &-0.1620\\
		&			&30		&{\bf 0.1741} &0.1021 &$0.1933^*$ &-0.2874\\	\hline	
			
	2.	&Squared	&10		&{\bf 0.3362} &0.2969 &0.2634 &-0.2096\\
		&Euclidean	&20		&{\bf 0.2469} &0.1522 &$0.3726^*$ &-0.5899\\
		&			&30		&{\bf 0.1658} &0.0440 &$0.2933^*$ &-0.6083\\	\hline
			
	3.	&City-block	&10		&{\bf 0.2369} &0.2354 &0.1703 &-0.2278\\
		&			&20		&{\bf 0.2019} &0.1870 &0.1879 &-0.2398\\
		&			&30		&{\bf 0.1752} &0.1524 &$0.1988^*$ &-0.2868\\	\hline
			
	4.	&Correlation&10		&{\bf 0.3367} &0.2560 &0.2582 &-0.0060\\
		&			&20		&{\bf 0.2291} &0.0899 &0.0867 &-0.4120\\
		&			&30		&{\bf 0.1742} &-0.0349 &0.0998 &-0.7018\\	
		\hline	
	\end{tabular}
	\caption{\label{sampling500}Silhouette Values for modified SC and HC with Pivotal Sampling and VQ for $N=500$. Silhouette Values in bold represent good clustering.}
\end{table}

When we compare our algorithm (values in the fourth column, and highlighted in bold) with other variants, it is evident that we are clearly better than modified SC with VQ and HC with VQ (values in the fifth and the seventh columns); our values are higher than those from these two algorithms.

When we compare our algorithm with HC with Pivotal Sampling (values in the sixth column), we again perform better for many cases. However, for some cases, our algorithm performs worse than HC with Pivotal Sampling (highlighted with a *). Upon further analysis (discussed below), we realize that an inappropriate grouping of genotypes by HC with Pivotal Sampling results in these set of Silhouette Values getting wrongly inflated.

\begin{figure}[ht]
\centering
\includegraphics[width=\linewidth,height=6cm]{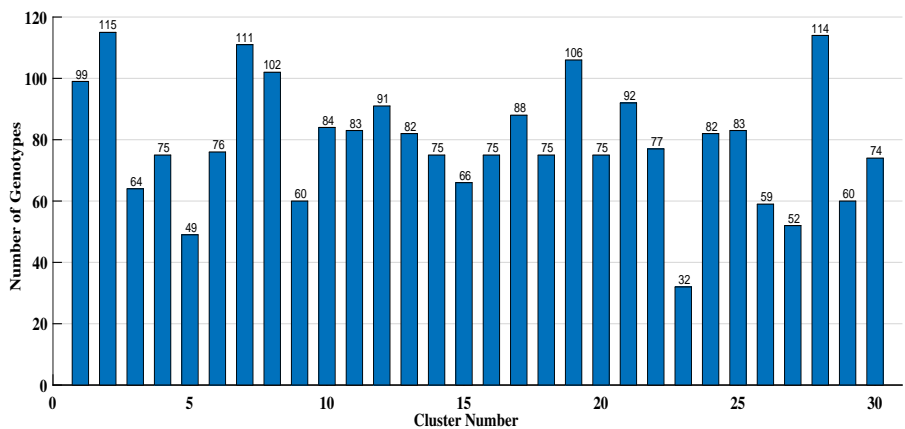}
	\caption{Distribution of Genotypes (modified SC with Pivotal Sampling) for Squared Euclidean similarity measure and cluster size thirty.}
	\label{Fig2}
\end{figure}

\begin{figure}[ht]
\centering
\includegraphics[width=\linewidth,height=6cm]{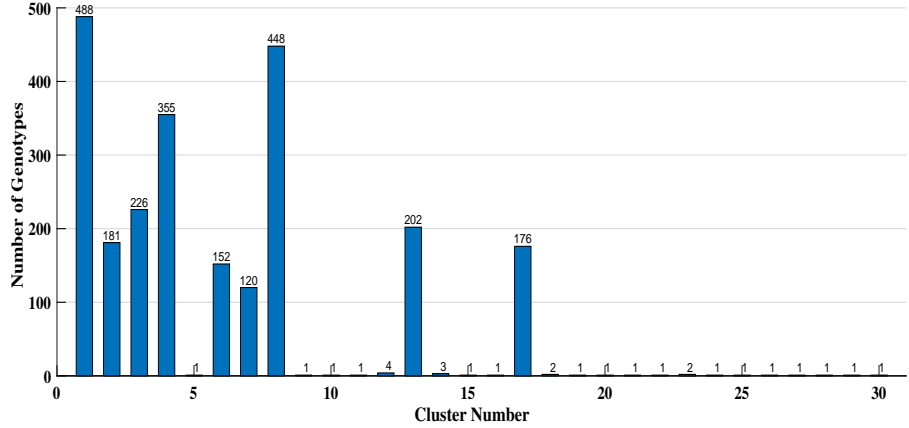}
	\caption{Distribution of Genotypes (HC with Pivotal Sampling) for Squared Euclidean similarity measure and cluster size thirty.}
	\label{Fig3}
\end{figure}

To further assess the quality of our algorithm, we present the distribution of genotypes into different clusters (after reverse-mapping) for modified SC with Pivotal Sampling and HC with Pivotal Sampling.
Without loss of generality, this comparison is done using the Squared Euclidean similarity measure and cluster size thirty.
The results for our algorithm are given in Figure \ref{Fig2} and for HC with Pivotal Sampling are given in Figure \ref{Fig3}. 
On the $x$-axis, we have the cluster number and on the $y$-axis, the number of genotypes present in them. 

From Figure \ref{Fig2}, we can see that our algorithm equally distributes all genotypes between the different clusters. This matches the real-life segregation as well since all genotypes belong to the same plant (Soybean in this case). This clustering has also been validated by the plant biologists at Indian Institute of Soybean Research, Indore, India.
The equivalent clustering results obtained by using HC with Pivotal Sampling (as given in Figure \ref{Fig3}), however, depicts a very skewed distribution. Most genotypes are segregated in only a few clusters, while the remaining clusters contain only one or two genotypes.

This is also the reason for the inflation of Silhouette Values of HC with Pivotal Sampling in Table \ref{sampling500} since the intra-cluster similarity for solitary genotype is zero leading to its respective Silhouette Value to become one (the maximum possible; see Eq. (\ref{eq:2})). Thus, our algorithm also outperforms HC with Pivotal Sampling, which from Table \ref{sampling500} was not very evident.

{\it Next}, as mentioned earlier, to further demonstrate the applicability of our work, we also present the results with a sample size $300$. Since modified SC with VQ turns out to be our closest competitor, we compare our algorithm with this one only. This comparison is given in Table \ref{sampling300}, with its columns mapping the respective columns of Table \ref{sampling500}. As evident from Table \ref{sampling300}, our modified SC with Pivotal Sampling substantially outperforms modified SC with VQ (see values in columns 4 and 5).

\begin{table}[ht]
\centering
	\begin{tabular}{|c|c|c|c|c|} \hline
			\textbf{Sr.}& \textbf{Similarity} &\textbf{\# of} & \multicolumn{2}{|c|}{\textbf{modified SC}}  \\
			\cline{4-5}
			
			\textbf{No.}	& \textbf{Measure}	& 	\textbf{Clusters}	& \textbf{Pivotal}	& \textbf{VQ}     \\ 
			
			& 		&  $(k)$	& \textbf{Sampling} 	&  	\\ \hline
			
			1.	&Euclidean	&10	&{\bf }0.2104	&0.1833 \\
			&				&20	&{\bf }0.1968	&0.0955 \\
			&				&30	&{\bf }0.1743	&0.0722 \\	\hline	
			
			2.	&Squared	&10	&{\bf }0.3280	&0.2589 \\
				&Euclidean	&20	&{\bf }0.2424	&0.1322 \\
			&				&30	&{\bf }0.1613	&0.0044 \\	\hline
			
			3.	&City-block	&10	&{\bf }0.2392	&0.2157 \\
			&				&20	&{\bf }0.1990	&0.1696 \\
			&				&30	&{\bf }0.1752	&0.1373 \\	\hline

			4.	&Correlation&10	&{\bf }0.3368	&0.2229	\\
			&				&20	&{\bf }0.2312	&0.0336	\\
			&				&30	&{\bf }0.1725	&-0.0788 \\	\hline		
			
	\end{tabular}
\caption{\label{sampling300}Silhouette Values for modified SC with Pivotal Sampling and VQ for $N=300$.}	
\end{table}

As earlier, {\it finally}, we compare the results of our algorithm (modified SC with Pivotal Sampling) with the currently popular clustering algorithm in the plant studies domain (i.e. HC without sampling). For this set of experiments, without loss of generality, we use the cluster size of ten. The results of this comparison are given in Table \ref{percentage}, where the first four columns are self-explanatory (based upon the data given in Tables \ref{sampling500} and \ref{sampling300} earlier). In the last column of this table, we also evaluate the percentage improvement in our algorithm over HC.
As evident from this table, our algorithm is up to 45\% more accurate than HC for both the sample sizes.
As earlier, our algorithm also has the crucial added benefit of reduced computational complexity as compared to HC.

\begin{table}[ht]
\centering
	\begin{tabular}{|c|c|c|c|c|} \hline
		\textbf{Sample}	& \textbf{Similarity} & \textbf{modified SC with}	& \textbf{HC}  & \textbf{Percentage}	   \\
			
		\textbf{Size}		& \textbf{Measure} 	& \textbf{Pivotal Sampling}  &  & \textbf{Improvement} \\	\hline
			
		& Euclidean	& 0.2152	& 0.2173 	& -0.97\% \\
	$N=500$	&Squared Euclidean 	& 0.3362 &0.3257 	& 3.22\%    \\
		&City-block	& 0.2369	& 0.2135	&  10.96\%  \\
		&Correlation	& 0.3367 & 0.2307	&  45.95\% \\	\hline
		
		& Euclidean	& 0.2104	& 0.2173 	& -3.28\% \\
	$N=300$	&Squared Euclidean 	& 0.3280 &0.3257 	& 0.71\%    \\
		&City-block	& 0.2392	& 0.2135	&   12.04\% \\
		&Correlation	& 0.3368 & 0.2307	&  45.99\% \\	\hline
			
	\end{tabular}
\caption{\label{percentage}Silhouette Values of modified SC with Pivotal Sampling and HC for cluster size ten.}	
\end{table}


\section{Conclusions and Future Work}
\label{sec:6}
We present the modified Spectral Clustering (SC) with Pivotal Sampling algorithm for clustering plant genotypes using their phenotypic data. We use SC for its accurate clustering and Pivotal Sampling for its effective sample selection that in-turn makes our algorithm scalable for large data. Since building the similarity matrix is crucial for the SC algorithm, we exhaustively adapt seven similarity measures to build such a matrix. We also present a novel way of assigning probabilities to different genotypes for Pivotal Sampling. 

We perform two sets of experiments on about 2400 Soybean genotypes that demonstrate the superiority of our algorithm. 
\textit{First}, when compared with the competitive clustering algorithms with samplings (SC with Vector Quantization (VQ), Hierarchical Clustering (HC) with Pivotal Sampling, and HC with VQ), Silhouette Values obtained when using our algorithm are higher. \textit{Second}, our algorithm doubly outperforms the standard HC algorithm in terms of clustering accuracy and computational complexity. We are up to 45\% more accurate and an order of magnitude faster than HC.

Since the choice of the similarity matrix has a significant impact on the quality of clusters, in the future, we intend to adapt other ways of constructing this matrix such as Pearson $\chi^2$, Squared $\chi^2$, Bhattacharyya, Kullback-Liebler etc. \citep{distance}.
Furthermore, in-place of Pivotal Sampling, we also plan to adapt other probabilistic sampling techniques like Cube Sampling, which possess complementary data analysis properties \citep{sampling}. 
As mentioned earlier, our algorithm is developed to work well for phenotypic data of all plants. Hence, in the future, we aim to test our algorithm on other plant genotypes as well (e.g., Wheat, Rice, etc.) \citep{wheat,rice-future}.

\section*{Acknowledgments}
The authors would like to thank Mr. Mohit Mohata, Mr. Ankit Gaur and Mr. Suryaveer Singh (IIT Indore, India) for their help in preliminary experiments, which they did as part of their undergraduate degree project. We would also like to sincerely thank Dr. Vangala Rajesh and Dr. Sanjay Gupta (Indian Institute of Soybean Research, Indore, India) for their help in generating the experimental data.


\begin{thebibliography}{0}
\bibitem{cite2} Painkra P, Shrivatava R, Nag SK and Markam NK, Clustering analysis of soybean germplasm ({\it Glycine max} L. Merrill), {\it The Pharma Innovation Journal}, {\bf 7(4)}:781--786, 2018.

\bibitem{Ingvarsson2011} Ingvarsson PK and Street NR, Association genetics of complex traits in plants, {\it New Phytologist}, {\bf 189(4)}:909--922, 2011.	
	
\bibitem{reseq302} Zhou Z, Jiang Y, Wang Z, Gou Z, Lyu J, Li W, Yu Y, Shu L, Zhao Y, Ma Y, Fang C, Shen Y, Liu T, Li C, Li Q, Wu M, Wang M, Wu Y, Dong Y, Wan W, Wang X, Ding Z, Gao Y, Xiang H, Zhu B, Lee S, Wang W and Tian Z, Resequencing 302 wild and cultivated accessions identifies genes related to domestication and improvement in soybean, {\it Nature Biotechnology}, {\bf 33(4)}:408--414, 2015.

\bibitem{sequencing} Subramanian S, Ramasamy U and Chen D, VCF2PopTree: a client-side software to construct population phylogeny from genome-wide SNPs, {\it PeerJ}, {\bf 7}:e8213, 2019.
	
\bibitem{wheat} Sharma P, Sareen S, Saini M, Verma A, Tyagi BS and Sharma I, Assessing genetic variation for heat tolerance in synthetic wheat lines using phenotypic data and molecular markers, {\it Australian Journal of Crop Science}, {\bf 8(4)}:515--522, 2014.

\bibitem{cluster-analysis-1} Kahraman A, Onder M and Ceyhan E, Cluster analysis in common bean genotypes ({\it Phaseolus vulgaris} L.), {\it Turkish Journal of Agricultural and Natural Sciences}, {\bf 1}:1030--1035, 2014.



\bibitem{hc} Rokach L, A survey of clustering algorithms, in Maimon O, Rokach L (eds.), {\it Data Mining and Knowledge Discovery Handbook}, Springer, Boston, MA, pp.~269--298, 2009.

\bibitem{sc} Luxburg UV, A tutorial on spectral clustering, {\it Statistics and Computing}, {\bf 17(4)}:395--416, 2007.

\bibitem{vqsc} Shastri AA, Ahuja K, Ratnaparkhe MB, Shah A, Gagrani A and Lal A, Vector quantized spectral clustering applied to whole genome sequences of plants, {\it Evolutionary Bioinformatics}, {\bf 15}:1--7, 2019.

\bibitem{hc_complexity} Mullner D, fastcluster: fast hierarchical, agglomerative clustering routines for R and Python, {\it Journal of Statistical Software}, {\bf 53(9)}:1--8, 2013.

\bibitem{sampling} Tille Y, Sampling algorithms, Springer-Verlag New York, 2006.

\bibitem{pivotal} Chauvet G, On a characterization of ordered pivotal sampling, {\it Bernoulli}, {\bf 18(4)}:1320--1340, 2012.

\bibitem{phenotypic-data} Gireesh C, Husain SM, Shivakumar M, Satpute GK, Kumawat G, Arya M, Agarwal DK and Bhatia VS, Integrating principal component score strategy with power core method for development of core collection in Indian soybean germplasm, {\it Plant Genetic Resources}, {\bf 15(3)}:230--238, 2015. 

\bibitem{rice1} Immanuel SC, Pothiraj N, Thiyagarajan K, Bharathi M and Rabindran R, Genetic parameters of variability, correlation and path-coefficient studies for grain yield and other yield attributes among rice blast disease resistant genotypes of rice ({\it Oryza sativa} L.), {\it African Journal of Biotechnology}, {\bf 10(17)}:3322--3334, 2011.

\bibitem{rice2} Divya B, Robin S, Biswas A and Joel JA, Genetics of association among yield and blast resistance traits in rice ({\it Oryza sativa}), {\it Indian Journal of Agricultural Sciences}, {\bf 85(3)}:354--360, 2015.

\bibitem{leaf} Huang F, Gan Y, Zhang D, Deng F and Peng J, Leaf shape variation and its correlation to phenotypic traits of Soybean in northeast China. {\it Proc. of the 6th Int. Conf. on Bioinformatics and Computational Biology}, Association for Computing Machinery, New York, pp.~40--45, 2018.

\bibitem{endobacteria} Carpentieri-Pipolo V, de Almeida Lopes KB and Degrassi G, Phenotypic and genotypic characterization of endophytic bacteria associated with transgenic and non-transgenic soybean plants, {\it Archives of Microbiology}, {\bf 201(8)}:1029--1045, 2019.

\bibitem{wheat2} Dodig D, Zoric M, Kobiljski B, Surlan-Momirovic G and Quarrie SA, Assessing drought tolerance and regional patterns of genetic diversity among spring and winter bread wheat using simple sequence repeats and phenotypic data, {\it Crop and Pasture Science}, {\bf 61(10)}:812--824, 2010.

\bibitem{corn} Stansluos AA, Ozturk A, Kodaz S, Pour AH and Sylvestre H, Genetic diversity in sweet corn ({\it Zea mays} L. {\it saccharata}) cultivars evaluated by agronomic traits, {\it Mysore Journal of Agricultural Sciences}, {\bf 53(1)}:1--8, 2019.

\bibitem{root} Fried HG, Narayanan S and Fallen B, Characterization of a soybean ({\it Glycine max} L. Merr.) germplasm collection for root traits, {\it PloS ONE}, {\bf 13(7)}:e0200463, 2018.

\bibitem{pivotal2} Deville JC and Tille Y, Unequal probability sampling without replacement through a splitting method, {\it Biometrika}, {\bf 85(1)}:89--101, 1998.

\bibitem{distance} Cha SH, Comprehensive survey on distance/similarity measures between probability density functions, {\it International Journal of Mathematical Models and Methods in Applied Sciences}, {\bf 4(1)}:300--307, 2007.	

\bibitem{correlation} Szekely GJ, Rizzo ML and Bakirov NK, Measuring and testing dependence by correlation of distances, {\it The Annals of Statistics}, {\bf 35(6)}:2769--2794, 2007.

\bibitem{hamming} Norouzi M, Fleet DJ and Salakhutdinov RR, Hamming distance metric learning, {\it Proc. of the 25th Int. Conf. on Advances in Neural Information Processing Systems}, pp.~1061--1069, 2012.

\bibitem{jaccard} Matlab Documentation, Pdist - pairwise distance between pairs of observations, available at \url{https://in.mathworks.com/help/stats/pdist.html}, (accessed 22 June 2020), 2006.

\bibitem{normalize} Jain A, Nandakumar K and Ross A, Score normalization in multimodal bio-metric systems, {\it Pattern Recognition}, {\bf 38(12)}:2270--2285, 2005.

\bibitem{normalize1} Shastri AA, Tamrakar D and Ahuja K, Density-wise two stage mammogram classification using texture exploiting descriptors, {\it Expert Systems with Applications}, {\bf 99}:71--82, 2018.

\bibitem{NgSC} Ng AY, Jordan MI and Weiss Y, On spectral clustering: analysis and an algorithm, {\it Proc. of the 15th Int. Conf. on Advances in Neural Information Processing Systems}, pp.~849--856, 2002.

\bibitem{spsc} Nemade V, Shastri AA, Ahuja K and Tiwari A, Scaled and projected spectral clustering with vector quantization for handling big data. {\it Proc. of the Symposium Series on Computational Intelligence (SSCI)}, pp.~2174--2179, 2018.

\bibitem{HT} Horvitz DG and Thompson DJ, A generalization of sampling without replacement
from a finite universe, {\it Journal of the American Statistical Association}, {\bf 47}:663--685, 1952.

\bibitem{hajek} H\'ajek J, Comment on ``An essay on the logical foundations of survey sampling, part one", in Godambe VP, Sprott DA (eds.), {\it Foundations of Statistical Inference}, Rinehart and Winston, Toronto, 1971.

\bibitem{validation} Fahad A, Alshatri N, Tari Z, Alamri A, Khalil I, Zomaya AY, Foufou S and Bouras A, A survey of clustering algorithms for big data: taxonomy and empirical analysis, {\it IEEE Transactions on Emerging Topics in Computing}, {\bf 2(3)}:267--279, 2014.

\bibitem{silhouette} Rousseeuw PJ, Silhouettes: a graphical aid to the interpretation and validation of cluster analysis, {\it Journal of Computational and Applied Mathematics}, {\bf 20}:53--65, 1987.

\bibitem{eigenheuristic} Kong W, Sun C, Hu S and Zhang J, Automatic spectral clustering and its application, {\it Proc. of the 3rd Int. Conf. on Intelligent Computation Technology and Automation}, IEEE, pp.~841--845, 2010.

\bibitem{rice-future} Kim B, Classifying {\it Oryza sativa} accessions into {\it Indica} and {\it Japonica} using logistic regression model with phenotypic data, {\it PeerJ}, {\bf 7}:e7259, 2019.

\end{thebibliography}

\end{document}